\documentclass{article}

\usepackage{PRIMEarxiv}

\usepackage[utf8]{inputenc} 
\usepackage[T1]{fontenc}    
\usepackage{hyperref}       
\usepackage{url}            
\usepackage{booktabs}       
\usepackage{amsfonts}       
\usepackage{nicefrac}       
\usepackage{microtype}      
\usepackage{lipsum}
\usepackage{fancyhdr}       
\usepackage{graphicx}       
\graphicspath{{media/}}     

\newcommand{\mycomment}[1]{}

\title{OpenFPL: An open-source forecasting method rivaling state-of-the-art Fantasy Premier League services}

\author{
  Daniel Groos \\
  Groos Analytics \\
}

\makeatletter
\let\@toptitlebar\relax
\let\@bottomtitlebar\relax
\makeatother

\begin{document}

\maketitle

\begin{abstract}

Fantasy Premier League engages the football community in selecting the Premier League players who will perform best from gameweek to gameweek. Access to accurate performance forecasts gives participants an edge over competitors by guiding expectations about player outcomes and reducing uncertainty in squad selection. However, high-accuracy forecasts are currently limited to commercial services whose inner workings are undisclosed and that rely on proprietary data. This paper aims to democratize access to highly accurate forecasts of player performance by presenting OpenFPL, an open-source Fantasy Premier League forecasting method developed exclusively from public data. Comprising position-specific ensemble models optimized on Fantasy Premier League and Understat data from four previous seasons (2020-21 to 2023-24), OpenFPL achieves accuracy comparable to a leading commercial service when tested prospectively on data from the 2024-25 season. OpenFPL also surpasses the commercial benchmark for high-return players ($>$ 2 points), which are most influential for rank gains. These findings hold across one-, two-, and three-gameweek forecast horizons, supporting long-term planning of transfers and strategies while also informing final-day decisions. Models and inference code are freely available at \url{https://github.com/daniegr/OpenFPL}.

\end{abstract}


\section{Introduction}

More than 11 million people play Fantasy Premier League (FPL), an online game that rewards accurate forecasting of football‑player performance \cite{fpl}. Participants construct squads within budget constraints and score points from real match events in the English Premier League. Accurate forecasting of performance, including goals, assists, and clean sheets, despite injuries, rotation risk, and fixture congestion, is thus the primary lever for competitive advantage in this large‑scale, data‑driven decision problem.

Following the innovative study of Matthews et al. \cite{matthews2012}, various attempts have been made to accurately forecast FPL player performance, most of which have followed in the footsteps of Stolyarov and Vasiliev \cite{stolyarov2017} using machine learning methods trained on historical data \cite{gupta2019, bonello2019, rajesh2022, bangdiwala2022, pokharel2022, lombu2024, venter2024, frees2024, tamimi2025}, including a series of thesis studies \cite{eilertsen2018, dykman2018, saifi2018, lindberg2020, ramdas2022, valouxis2023}. Furthermore, there exists a growing pool of commercial FPL forecasting services \cite{fix, hub, scout, fplreview, transfer}. However, state-of-the-art methods use proprietary data sources such as expert minute projections and bookmaker feeds, and have not openly released their code or trained models, but instead lock their highest‑fidelity projections behind paywalls. The widely used Massive Data Model from FPL Review, for example, is accessible only via a subscription tier \cite{fplreview}. This barrier impedes reproducibility, hinders independent benchmarking, and excludes much of the global player base from high‑quality decision support. Consequently, the wider FPL community still lacks a transparent, high‑accuracy method built solely on public data.

\newpage
The present study aims to bridge this gap with OpenFPL, an open‑source forecasting method competing with the predictive accuracy of the commercial FPL Review Massive Data Model when prospectively tested on 2024‑25 gameweeks across multiple forecast horizons (i.e., one, two, and three gameweeks ahead). OpenFPL constitutes one ensemble regressor per FPL position (goalkeepers, defenders, midfielders, forwards, and assistant managers), each aggregating FPL point forecasts from XGBoost and Random Forest machine learning models, whose hyperparameters are tuned via automatic \textit{K}‑Best Search over cross‑validation folds. The models rely exclusively on publicly available data from the official FPL and Understat APIs \cite{fplapi, understatapi}. Importantly, OpenFPL dispenses with proprietary “expected minutes” projections, using instead the categorical availability tags provided by the FPL API. By demonstrating that an openly reproducible ensemble can compete with leading commercial systems, the aim is to democratize access to high‑quality FPL forecasts and provide a fully transparent baseline for researchers and practitioners. All trained models and inference code are available under an MIT license on GitHub\footnote{\url{https://github.com/daniegr/OpenFPL}}.

The main contributions are as follows:
\begin{itemize}
    \item \textbf{Public‑data forecasting.} OpenFPL attains parity with leading subscription‑based services while relying solely on freely available FPL and Understat data, eliminating dependence on proprietary minute projections, betting odds, or premium event feeds.
    \item \textbf{Position‑specific ensemble design.} Separate XGBoost and Random Forest ensembles are optimized and trained for each FPL position, including the recently introduced assistant managers, through automatic hyperparameter search.
    \item \textbf{Prospective multi‑horizon evaluation.} OpenFPL demonstrates state‑of‑the‑art accuracy at one-, two-, and three-gameweek horizons on out‑of‑sample 2024‑25 data, delivering superior forecasts for high‑return players who are key prospects for FPL participants.
    \item \textbf{Fully open research artifacts.} Optimized models and trained weights are released under an MIT license, providing a transparent benchmark for future FPL analytics research.
\end{itemize}

\section{Methods}

\begin{figure}
  \centering
  \includegraphics[width=\linewidth]{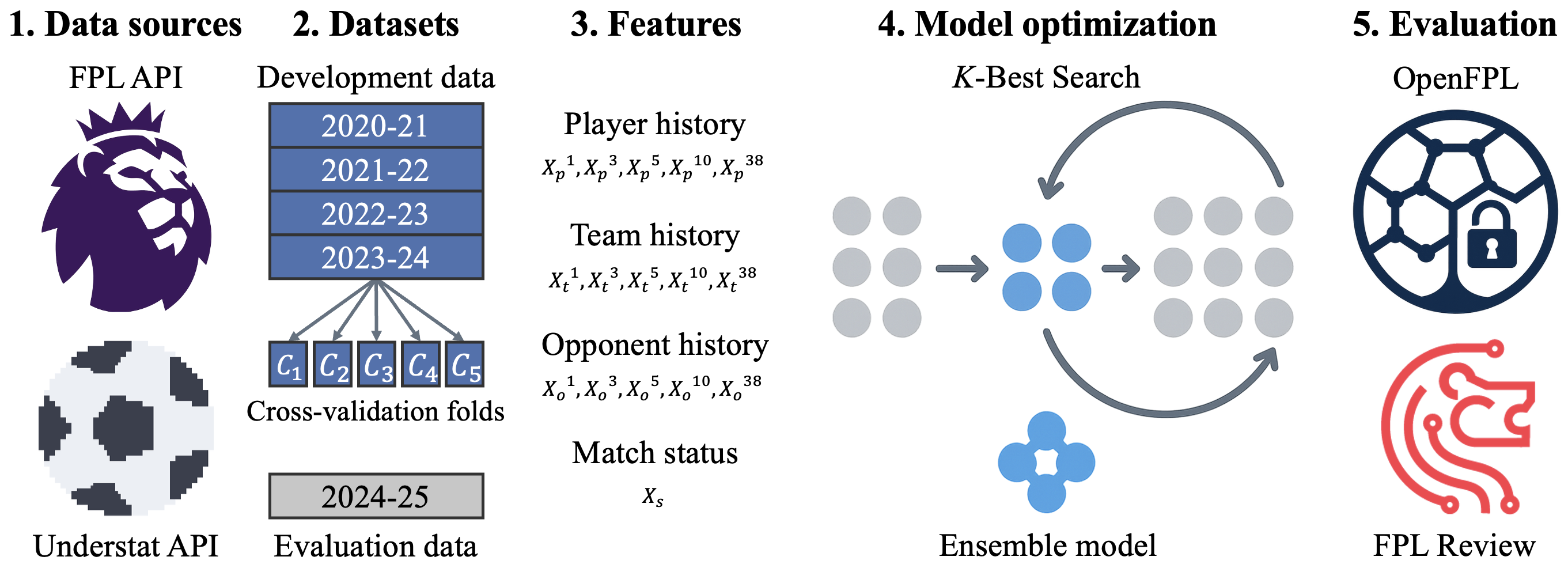}
  \caption{Overview of steps involved in developing and evaluating OpenFPL}
  \label{fig:openfpl}
\end{figure}

Figure \ref{fig:openfpl} depicts the overall process of developing and evaluating OpenFPL for high-accuracy forecasting of FPL player performance. The steps include 1) accessing open-source FPL and Understat data, 2) splitting data into datasets for development and evaluation, 3) defining features for machine learning models, 4) optimizing models and constructing ensembles, and 5) performing evaluation against state-of-the-art FPL service. Each step is described in detail in the respective manuscript section.

\newpage
\subsection{Data Sources}

All data used for developing and evaluating OpenFPL originate from the publicly available FPL and Understat APIs \cite{fplapi,understatapi}. The FPL API provides comprehensive FPL-related data for each gameweek, including player-level information such as position, availability, minutes played, goals, assists, saves, and FPL points. It also includes team-level information like fixture opponent and home/away status. In contrast, the Understat API offers advanced performance statistics at both player and team levels. Player-level metrics include expected goals (xG), expected goals against (xGA), expected assists (xA), shots, and key passes. At the team level, it provides indicators such as Deep and Passes Per Defensive Action (PPDA), which help quantify territorial dominance and pressing intensity. In the present study, historical FPL and Understat data were obtained from the FPL Historical Dataset \cite{vaastav}.

\subsection{Datasets}

FPL and Understat data from four consecutive seasons, from 2020-21 to 2023-24, were used for developing OpenFPL\footnote{Due to lack of Understat data for players in the 2020-21 season, 1-Nearest Neighbor was used to fill missing values by taking the most similar sample from the remaining three seasons of the development data (i.e., 2021-22, 2022-23, and 2023-24), under assumption of minutes played and same number of assists and goals scored.}, whereas gameweek data from the 2024-25 season were isolated for evaluation. The development data were further split into five cross-validation folds (i.e., $C_1$ to  $C_5$), with each of the 26 Premier League teams allocated to one of the folds. To ensure the individual partitions represent a mixture of high- and low-performing players, every cross-validation fold comprises both teams from the upper and lower half of the Premier League table. Furthermore, to achieve similar size of the cross-validation folds, teams appearing in all four seasons were combined with teams present in a subset of the seasons due to relegation or promotion. Consequently, each cross-validation fold constitutes a total of 16 team seasons. The overview of teams allocated to the five cross-validation folds of the development data is shown in Table \ref{tab:split}. For the evaluation dataset to represent a strict test of OpenFPL's ability to generalize to future FPL seasons, data from the 2024-25 season were not gathered until after completing development. Consequently, the evaluation dataset comprises prospective data of all 20 teams in a subset of the gameweeks (i.e., 32-38) from the 2024-25 season.

\begin{table}
\centering
\caption{The development data were divided into five cross-validation folds through split on Premier League teams}
\begin{tabular}{ll}
\toprule
\textbf{Cross-validation fold} & \textbf{Teams (Number of seasons)} \\
\midrule
$C_1$ & Everton (4), Leeds (3), Luton (1), Manchester United (4), West Ham (4) \\
$C_2$ & Aston Villa (4), Burnley (3), Leicester (3), Tottenham (4), Watford (1), West Bromwich (1) \\
$C_3$ & Bournemouth (2), Brentford (3), Manchester City (4), Southampton (3), Wolverhampton (4) \\
$C_4$ & Brighton (4), Chelsea (4), Crystal Palace (4), Nottingham Forest (2), Sheffield United (2) \\
$C_5$ & Arsenal (4), Fulham (3), Liverpool (4), Newcastle (4), Norwich (1) \\
\bottomrule
\end{tabular}
\label{tab:split}
\end{table}

\subsection{Features}

For machine learning models to accurately forecast FPL player performance, the models' input features should cover historical as well as current information relevant for the prediction. OpenFPL tackles this by utilizing features over different time horizons, from short-term (i.e., one, three, and five most recent matches) to long-term (i.e., past 10 and 38 matches), together with current data on the match status of players. Moreover, since there is variation in information that matters for different player positions (e.g., goals scored is less relevant for a goalkeeper and number of minutes can be neglected for an assistant manager), three feature sets are defined for goalkeepers (GK), defenders/midfielders/forwards (DEF/MID/FWD), and assistant managers (AM), respectively. The feature sets combine player-specific ($X_p$), team-specific ($X_t$), and opponent-specific ($X_o$) historical features, that are averaged over the aforementioned horizons. Additionally, match status ($X_s$) is represented for all positions, using the FPL categories for current availability of a player in percentage (i.e., 0\%, 25\%, 50\%, 75\%, or 100\%) for GK, DEF, MID, and FWD, and current league rank of team and opponent for AM. As shown in Table \ref{tab:features}, all positions harness a combination of features from the FPL and Understat APIs. 

\begin{table}
\centering
\caption{Feature sets for machine learning models of the different player positions (i.e., GK, DEF, MID, FWD, and AM). $X_p$, $X_t$, and $X_o$ are historical player, team, and opponent features, respectively, with each feature represented five times according to mean aggregation over one-, three-, five-, 10-, and 38-match horizons. $X_s$ defines the current status of the player before the match. $^*$Relevant FPL points are FPL points achieved by the player at the venue of the upcoming match (i.e., home or away). $^\dag$BPS = FPL Bonus Points System. $^\ddag$League rank is derived from the goals scored and conceded of all teams in the previous matches of the season. $^\mathsection$PPDA = Passes Per Defensive Action.}
\begin{tabular}{lllccc}
\toprule
\textbf{Feature group} & \textbf{Data source} & \textbf{Feature} & \textbf{GK} & \textbf{DEF/MID/FWD} & \textbf{AM} \\
\midrule
$X_p$ & FPL API & FPL points & \checkmark & \checkmark & \checkmark \\
&  & Relevant FPL points$^*$ & \checkmark & \checkmark & \checkmark \\
&  & Minutes played & \checkmark & \checkmark & \\
&  & Influence & \checkmark & \checkmark & \\
&  & Creativity & \checkmark & \checkmark & \\
&  & Threat & \checkmark & \checkmark & \\
&  & Goals scored & & \checkmark & \\
&  & Penalties missed & & \checkmark & \\
&  & Assists & \checkmark & \checkmark & \\
&  & Goals conceded & \checkmark & \checkmark & \\
&  & Own goals & \checkmark & \checkmark & \\
&  & Saves & \checkmark & & \\
&  & Penalties saved & \checkmark & & \\
&  & Yellow cards & \checkmark & \checkmark & \\
&  & Red cards & \checkmark & \checkmark & \\
&  & BPS$^\dag$ & \checkmark & \checkmark & \\
&  & FPL bonus points & \checkmark & \checkmark & \\
& Understat API & Shots & & \checkmark & \\
&  & xG & & \checkmark & \\
&  & xGChain & \checkmark & \checkmark & \\
&  & xGBuildup & \checkmark & \checkmark & \\
&  & Key passes & \checkmark & \checkmark & \\
&  & xA & \checkmark & \checkmark & \\
\midrule
$X_t$ & FPL API & Goals scored & \checkmark & \checkmark & \checkmark \\
 &  & Goals conceded & \checkmark & \checkmark & \checkmark \\
 &  & League rank$^\ddag$ &  &  & \checkmark \\
 &  & Opponent league rank$^\ddag$ &  &  & \checkmark \\
 & Understat API & xG & \checkmark & \checkmark & \checkmark \\
 &  & Deep allowed & \checkmark & \checkmark & \checkmark \\
 &  & PPDA allowed att$^\mathsection$ & \checkmark & \checkmark & \checkmark \\
 &  & PPDA allowed def$^\mathsection$ & \checkmark & \checkmark & \checkmark \\
 &  & xGA & \checkmark & \checkmark & \checkmark \\
 &  & Deep & \checkmark & \checkmark & \checkmark \\
 &  & PPDA att$^\mathsection$ & \checkmark & \checkmark & \checkmark \\
 &  & PPDA def$^\mathsection$ & \checkmark & \checkmark & \checkmark \\
\midrule
$X_o$ & FPL API & Goals scored & \checkmark & \checkmark & \checkmark \\
 &  & Goals conceded & \checkmark & \checkmark & \checkmark \\
 & Understat API & xG & \checkmark & \checkmark & \checkmark \\
 &  & Deep allowed & \checkmark & \checkmark & \checkmark \\
 &  & PPDA allowed att$^\mathsection$ & \checkmark & \checkmark & \checkmark \\
 &  & PPDA allowed def$^\mathsection$ & \checkmark & \checkmark & \checkmark \\
 &  & xGA & \checkmark & \checkmark & \checkmark \\
 &  & Deep & \checkmark & \checkmark & \checkmark \\
 &  & PPDA att$^\mathsection$ & \checkmark & \checkmark & \checkmark \\
 &  & PPDA def$^\mathsection$ & \checkmark & \checkmark & \checkmark \\
\midrule
$X_s$ & FPL API & Player availability & \checkmark & \checkmark & \\
 &  & Team league rank$^\ddag$ &  &  & \checkmark \\
 &  & Opponent league rank$^\ddag$ &  &  & \checkmark \\
\midrule
 &  & Number of features & 196 & 206 & 122 \\
\bottomrule
\end{tabular}
\label{tab:features}
\end{table}

\newpage
\subsection{Model Optimization}

By leveraging the proposed features on development data, position-specific regression models were optimized for forecasting FPL points through hyperparameter search and ensemble modeling. Hyperparameter search was performed on individual cross-validation folds for each position, using all player-match pairs for that position as training and validation samples. The features of the training and validation samples were normalized to the [0, 1] range. The target variable for each sample was the player's normalized FPL points in the upcoming match. To balance overall forecasting accuracy with the ability to predict high-performing players, those of greatest interest to FPL participants, weighting of training and validation samples was performed using position-specific discretization based on the entropy of the target distributions (i.e., 2, 3, 4, 3, and 5 bins were used for GK, DEF, MID, FWD, and AM, respectively). During hyperparameter search, variations of Random Forest and XGBoost regression models \cite{randomforest, xgboost} were examined, as specified by the search space in Table \ref{tab:searchspace}. The machine learning models were optimized for root mean square error (RMSE) by employing \textit{K}-Best Search with population size $\textit{K} = 10$ \cite{kbestsearch, kbestsearchgithub}. In the present study, the vanilla \textit{K}-Best Search was extended by introducing a preliminary search, to automatically determine the performance threshold as the lowest RMSE of the first \textit{K} candidates. For each player position, an ensemble model was constructed by combining the top \textit{K} models from the \textit{K}-Best Search of each cross-validation fold. Ensemble model forecasts are obtained as the median forecasted FPL points of the 50 individual models.

All code was implemented in Python. The scikit-learn library was used for data preprocessing, with feature and target normalization by \texttt{MinMaxScaler}, and sample weighting using \texttt{KBinsDiscretizer} with $encode = ``ordinal"$, followed by \texttt{compute\_sample\_weight} with $class\_weight = ``balanced"$, clipping at the 95th percentile, and rescaling to unit mean. For defining, training, and predicting with Random Forest and XGBoost models, the scikit-learn and xgboost libraries were used, respectively \cite{sklearn, xgboostlibrary}. Models were initialized with $random\_state = 42$ and $n\_jobs = -1$, and XGBoost employed $early\_stopping\_rounds = 30$ with $eval\_metric = ``rmse"$.

\begin{table}
\centering
\caption{The search space used by \textit{K}-Best Search, where a search candidate consists of a machine learning method, either Random Forest or XGBoost, along with choices for hyperparameters of the respective method}
\begin{tabular}{lll}
\toprule
\textbf{Machine learning method} & \textbf{Hyperparameter} & \textbf{Alternatives} \\
\midrule
Random Forest & $n\_estimators$ & $200, 400, 800$ \\
 & $max\_depth$ & $10, 20, None$ \\
 & $min\_samples\_split$ & $2, 5$ \\
 & $min\_samples\_leaf$ & $1, 2, 5$ \\
 & $max\_features$ & $``sqrt", 0.2, 1.0$ \\
 & $bootstrap$ & $True, False$ \\
\midrule
XGBoost & $n\_estimators$ & $300, 600, 1200$ \\
 & $max\_depth$ & $3, 5, 7$ \\
 & $learning\_rate$ & $0.01, 0.05, 0.1$ \\
 & $subsample$ & $0.5, 0.75, 1.0$ \\
 & $colsample\_bytree$ & $0.5, 0.75, 1.0$ \\
 & $min\_child\_weight$ & $1.0, 5.0$ \\
 & $gamma$ & $0.0, 0.1$ \\
 & $reg\_lambda$ & $1.0, 5.0$ \\
\bottomrule
\end{tabular}
\label{tab:searchspace}
\end{table}

\newpage
\subsection{Evaluation}

After completing OpenFPL development, prospective accuracy evaluation on 2024-25 gameweeks was performed. OpenFPL predictions were generated using up-to-date historical data and player match status retrieved from the FPL and Understat APIs on the day before each gameweek deadline. For fair comparison with a state-of-the-art FPL forecasting method, concurrent predictions from the FPL Review Massive Data Model were obtained via a paid subscription \cite{fplreview}. The commercial FPL Review service has previously demonstrated superior accuracy compared to other leading FPL services \cite{valouxis2023}. Additionally, as a baseline, forecasting based on a player's mean points from the last five matches (Last 5) was included. Accuracy was evaluated at one-, two-, and three-gameweek horizons (i.e., gameweeks ahead), using RMSE as the primary metric and mean absolute error (MAE) as a secondary metric. Furthermore, accuracy by player position (i.e., GK, DEF, MID, FWD, and AM) was assessed at one-gameweek horizon. To differentiate forecasting accuracy across low- and high-performing FPL players, evaluation samples were organized into four return categories:
\begin{itemize}
  \item \textit{Zeros}: Players who did not play and thus received zero FPL points
  \item \textit{Blanks}: Players who played but earned a maximum of two FPL points
  \item \textit{Tickers}: Players who returned three or four FPL points
  \item \textit{Haulers}: Players who achieved at least five FPL points
\end{itemize}

\section{Results}

Table \ref{tab:overall} shows that both OpenFPL and the FPL Review Massive Data Model consistently outperform the baseline method, reducing RMSE by 5-34\% across all four return categories. For the low-return categories, Zeros (non-playing) and Blanks ($\leq$ 2 FPL points), the commercial FPL Review method achieves the lowest RMSE and MAE. In contrast, OpenFPL is most accurate in the high-return categories, Tickers (3–4 FPL points) and Haulers ($\geq$ 5 FPL points). This trend holds across all three forecast horizons. For all methods, shorter horizon improves forecasting for low-return categories, especially for Zeros where one-gameweek-ahead predictions lower RMSE by 15-25\% relative to forecasts three gameweeks ahead. No systematic horizon effect is observed for the high-return categories.

OpenFPL and the FPL Review Massive Data Model also demonstrate superior accuracy to the Last 5 baseline for all five player positions (Table \ref{tab:positions}). No single method, however, dominates every return category within any position. Consistent with the overall results, the FPL Review method provides the lowest RMSE for non-playing samples (Zeros) in every position. For the remaining return categories, there is individual variation across positions. Notably, for Tickers among assistant managers, FPL Review has 26\% lower RMSE than OpenFPL, whereas OpenFPL demonstrates 19\% improvement in RMSE for Blanks among forwards relative to the FPL Review Massive Data Model.

\begin{table}
\centering
\caption{Overall accuracy of methods in RMSE (MAE) with multi-horizon evaluation (i.e., one, two, and three gameweeks ahead) across the four return categories (i.e., Zeros, Blanks, Tickers, and Haulers)}
\begin{tabular}{llllll}
\toprule
\textbf{Gameweeks ahead} & \textbf{Method} & \textbf{Zeros} & \textbf{Blanks} & \textbf{Tickers} & \textbf{Haulers} \\
\midrule
1 & Last 5 & 0.791 (0.270) & 1.400 (0.652) & 2.136 (1.645) & 5.613 (4.709) \\
 & FPL Review & \textbf{0.689} (0.237) & \textbf{1.189} (0.597) & 1.594 (1.227) & 5.172 (4.381) \\
 & OpenFPL & 0.818 (0.427) & 1.291 (0.749) & \textbf{1.517} (1.127) & \textbf{5.142} (4.317) \\
\midrule
2 & Last 5 & 0.918 (0.315) & 1.372 (0.665) & 2.365 (1.711) & 5.573 (4.726) \\
 & FPL Review & \textbf{0.826} (0.313) & \textbf{1.219} (0.631) & 1.605 (1.228) & 5.169 (4.447) \\
 & OpenFPL & 0.922 (0.475) & 1.309 (0.765) & \textbf{1.569} (1.192) & \textbf{5.051} (4.289) \\
\midrule
3 & Last 5 & 0.971 (0.348) & 1.379 (0.683) & 2.002 (1.562) & 5.670 (4.851) \\
 & FPL Review & \textbf{0.918} (0.372) & \textbf{1.260} (0.666) & 1.402 (1.099) & 5.197 (4.512) \\
 & OpenFPL & 0.966 (0.506) & 1.306 (0.775) & \textbf{1.369} (1.079) & \textbf{5.171} (4.467) \\
\bottomrule
\end{tabular}
\label{tab:overall}
\end{table}

\newpage
\begin{table}
\centering
\caption{Accuracy of methods in RMSE (MAE) by player position (i.e., GK, DEF, MID, FWD, and AM) at one-gameweek horizon over the four return categories (i.e., Zeros, Blanks, Tickers, and Haulers)}
\begin{tabular}{llllll}
\toprule
\textbf{Position} & \textbf{Method} & \textbf{Zeros} & \textbf{Blanks} & \textbf{Tickers} & \textbf{Haulers} \\
\midrule
GK & Last 5 & 0.672 (0.155) & 1.110 (0.410)	& 1.533 (1.280) & 5.711 (4.930) \\
 & FPL Review & \textbf{0.512} (0.103) & 0.918 (0.369) & \textbf{1.150} (0.834) & \textbf{5.040} (4.335) \\
 & OpenFPL & 0.616 (0.208) & \textbf{0.888} (0.406) & 1.180 (0.807) & 5.678 (4.960) \\
\midrule
DEF & Last 5 & 0.753 (0.283) & 1.353 (0.667) & 2.818 (2.300) & 5.384 (4.735) \\
 & FPL Review & \textbf{0.740} (0.301) & \textbf{1.118} (0.604) & 1.544 (1.409) & \textbf{5.016} (4.459) \\
 & OpenFPL & 0.812 (0.482) & 1.129 (0.723) & \textbf{1.448} (1.223) & 5.062 (4.505) \\
\midrule
MID & Last 5 & 0.831 (0.292) & 1.176 (0.593) & 2.009 (1.593) & 5.788 (4.788) \\
 & FPL Review & \textbf{0.686} (0.235) & \textbf{1.000} (0.529) & 1.531 (1.197) & 5.559 (4.592) \\
 & OpenFPL & 0.902 (0.454) & 1.189 (0.744) & \textbf{1.375} (1.020) & \textbf{5.274} (4.235) \\
\midrule
FWD & Last 5 & 0.877 (0.283) & 1.321 (0.622) & 3.533 (5.013) & 5.586 (4.808) \\
 & FPL Review & \textbf{0.709} (0.201) & 1.261 (0.636) & 3.822 (3.047) & \textbf{4.621} (4.009) \\
 & OpenFPL & 0.719 (0.410) & \textbf{1.024} (0.646) & \textbf{2.694} (2.266) & 5.235 (4.722) \\
\midrule
AM & Last 5 & N/A & 6.116 (5.065) & 2.682 (1.927) & 5.642 (4.260) \\
 & FPL Review & N/A & \textbf{5.107} (4.796) & \textbf{2.183} (1.810) & 4.906 (3.922) \\
 & OpenFPL & N/A & 6.192 (6.076) & 2.937 (2.847) & \textbf{4.598} (3.471) \\
\bottomrule
\end{tabular}
\label{tab:positions}
\end{table}

\newpage
\section{Discussion}

The main objective of the study was to propose an open-source, public-data forecasting method competing with state-of-the-art commercial FPL services. In prospective accuracy evaluation, the OpenFPL method using solely openly available FPL and Understat data demonstrated accuracy non-inferior to a cutting-edge commercial method. On high-return categories, OpenFPL even improved accuracy by up to 5\% across player positions. 

The high accuracy of OpenFPL was achieved through position-specific ensemble models that were optimized directly for FPL point regression. This stands in contrast to acknowledged FPL forecasting methods harnessing an indirect approach forecasting FPL points from likelihood of rewarding events (e.g., goals, assists, and saves) \cite{fplreview,fplcopilot}. By use of \textit{K}-Best Search, hyperparameters of XGBoost and Random Forest regression models in OpenFPL were automatically tuned to accurately forecast FPL points from proposed feature sets, including historical player, team, and opponent data of different time horizons as well as up-to-date match status of players. Future research could assess whether an indirect public-data forecasting method employing hyperparameter search (e.g., \textit{K}-Best Search or Optuna \cite{kbestsearch,akiba2019optuna}) and ensemble modeling could improve accuracy further. It would also be valuable to systematically investigate for individual player positions the importance of different features and time horizons of historical features, to understand what features drive accurate predictions and whether accuracy gains can be achieved with other feature sets. A step in this direction would be to extend the search space with hyperparameters related to input features.

OpenFPL's superior accuracy in high-return players offers tangible benefits to FPL participants. By more reliably forecasting Tickers and Haulers, the players earning more than two FPL points (i.e., the standard reward for 60+ minutes played), OpenFPL can potentially drive rank gains. Participants can target differential picks or captaincy candidates with a higher expected upside, and they can acquire emerging assets before widespread ownership, thereby increasing differential value and reducing transfer-hit costs. Furthermore, OpenFPL’s high-return forecasts can inform price-change models such as Khamsan and Maskat \cite{khamsan2019handling}, improving the prediction of market movements and helping participants capture value rises ahead of competitors.

The proprietary FPL Review method retains an edge in forecasting the low-return players, Zeros and Blanks. These two categories are also the only ones that exhibit a systematic horizon effect. Interestingly, the horizon effect is stronger for the FPL Review Massive Data Model than for OpenFPL. This pattern likely reflects the importance of precise information on player availability and expected minutes when predicting low returns. Although the FPL API encodes injuries and suspensions, it does not indicate whether a player is expected to start a match, come on as a substitute, or be rested ahead of more important fixtures. FPL Review and other commercial methods, such as the Transfer Algorithm and FPL Copilot \cite{transfer,fplcopilot}, integrate detailed expected-minutes projections derived from team news, fixture congestion, and betting odds. Recent studies have explored incorporating such information in machine learning methods \cite{bonello2019,frees2024,tamimi2025}, and future work could extend this line through crowd-sourced minutes forecasts and web-scraping AI agents \cite{he2024webvoyager,ayuso2024manual}.

OpenFPL point forecasts can be integrated directly into existing team-selection frameworks, transforming predictive accuracy into actionable transfer, chip, and captaincy decisions. Previous studies have framed the week-to-week squad-building challenge as a constrained optimization task that embeds FPL rules. The majority of approaches have based optimization on linear programming \cite{gupta2019,rajesh2022,venter2024}, although evolutionary and multi-criteria methods have also been explored \cite{aribo2024machine,shah2023multi}. The widely used, open-source FPL Optimization Repository \cite{sertalp} provides a linear programming workflow and therefore offers a natural starting point for coupling team selection with OpenFPL. Replacing its default FPL Review inputs with OpenFPL forecasts would create the first fully open-source end-to-end system that combines state-of-the-art FPL forecasting with transparent optimization. This would also facilitate establishing an ecosystem that respond swiftly to FPL changes, as scoring rules, chip mechanics, and player classifications evolve almost every season.

\section{Conclusion}

OpenFPL, a fully open-source Fantasy Premier League forecasting method developed exclusively from publicly available data, delivers overall accuracy comparable to the leading commercial service and surpasses it for the high-return players that drive rank gains. Conversely, the proprietary method retains an edge on low-return outcomes. Because OpenFPL is transparent, reproducible, and free, its forecasts can be plugged directly into optimization-based team-selection and price-change pipelines and rapidly adapted to changes in FPL rules, chip mechanics, and player classifications. Taken together, these findings indicate that a community-maintained, open ecosystem can rival commercial FPL services precisely where it matters most to game participants, while lowering barriers to research and fostering further methodological development.

\bibliographystyle{unsrt}  
\bibliography{references}  

\end{document}